\documentclass[11pt,a4paper]{article}
\usepackage[hyperref]{emnlp2020}
\usepackage{times}
\usepackage{graphicx}
\usepackage{xcolor}
\usepackage{latexsym}
\usepackage{booktabs}

\definecolor{tblue}{RGB}{93, 142, 150}

\usepackage{microtype}

\aclfinalcopy 
\def\aclpaperid{13} 

\title{An Ensemble Approach for Automatic Structuring of Radiology Reports}

\author{Morteza Pourreza Shahri\textsuperscript{1} \\
  CodaMetrix  \\\And
  Amir Tahmasebi\textsuperscript{1} \\
  CodaMetrix\\ \And
  Bingyang Ye\textsuperscript{1} \\
  CodaMetrix  \\\AND
  Henghui Zhu\textsuperscript{2} \\
  Amazon \\\And
  Javed Aslam\textsuperscript{1} \\
  CodaMetrix\\\\
  \textsuperscript{1}\texttt{\{morteza,amir,bingyang,jay\}@codametrix.com}  \\
  \textsuperscript{2}\texttt{henghui@amazon.com} \\
  \textsuperscript{3}\texttt{ferris.timothy@mgh.harvard.edu} \\ \And
  Timothy Ferris\textsuperscript{3} \\
  Mass General Physicians Organization \\ \\
  
  }

\date{}

\begin{document}
\maketitle
\begin{abstract}
Automatic structuring of electronic medical records is of high demand for clinical workflow solutions to facilitate extraction, storage, and querying of patient care information. However, developing a scalable solution is extremely challenging, specifically for radiology reports, as most healthcare institutes use either no template or department/institute specific templates. Moreover, radiologists' reporting style varies from one to another as sentences are telegraphic and do not follow general English grammar rules. We present an ensemble method that consolidates the predictions of three models, capturing various attributes of textual information for automatic labeling of sentences with section labels. These three models are: 1) Focus Sentence model, capturing context of the target sentence; 2) Surrounding Context model, capturing the neighboring context of the target sentence; and finally, 3) Formatting/Layout model, aimed at learning report formatting cues. We utilize Bi-directional LSTMs, followed by sentence encoders, to acquire the context. Furthermore, we define several features that incorporate the structure of reports. We compare our proposed approach against multiple baselines and state-of-the-art approaches on a proprietary dataset as well as 100 manually annotated radiology notes from the MIMIC-III dataset, which we are making publicly available. Our proposed approach significantly outperforms other approaches by achieving 97.1\% accuracy. 
\end{abstract}

\section{Introduction}

Electronic medical records~(EMRs), such as radiology reports, contain patient clinical information and are often in the form of ``natural language" written or transcribed by providers~\citep{denny2008development}. Gathering and disseminating patient information from such notes is required for patient care management. Natural Language Processing (NLP)-driven solutions have been proposed to augment clinical workflows to facilitate such information extraction and structuring processes. Segmentation of medical reports into topically cohesive sections~\citep{cho2003automatic} is essential for NLP tasks such as relation extraction, Named Entity Recognition (NER), and Question and Answering.

\begin{figure}[htbp]
    \centering
    \includegraphics[width=\linewidth]{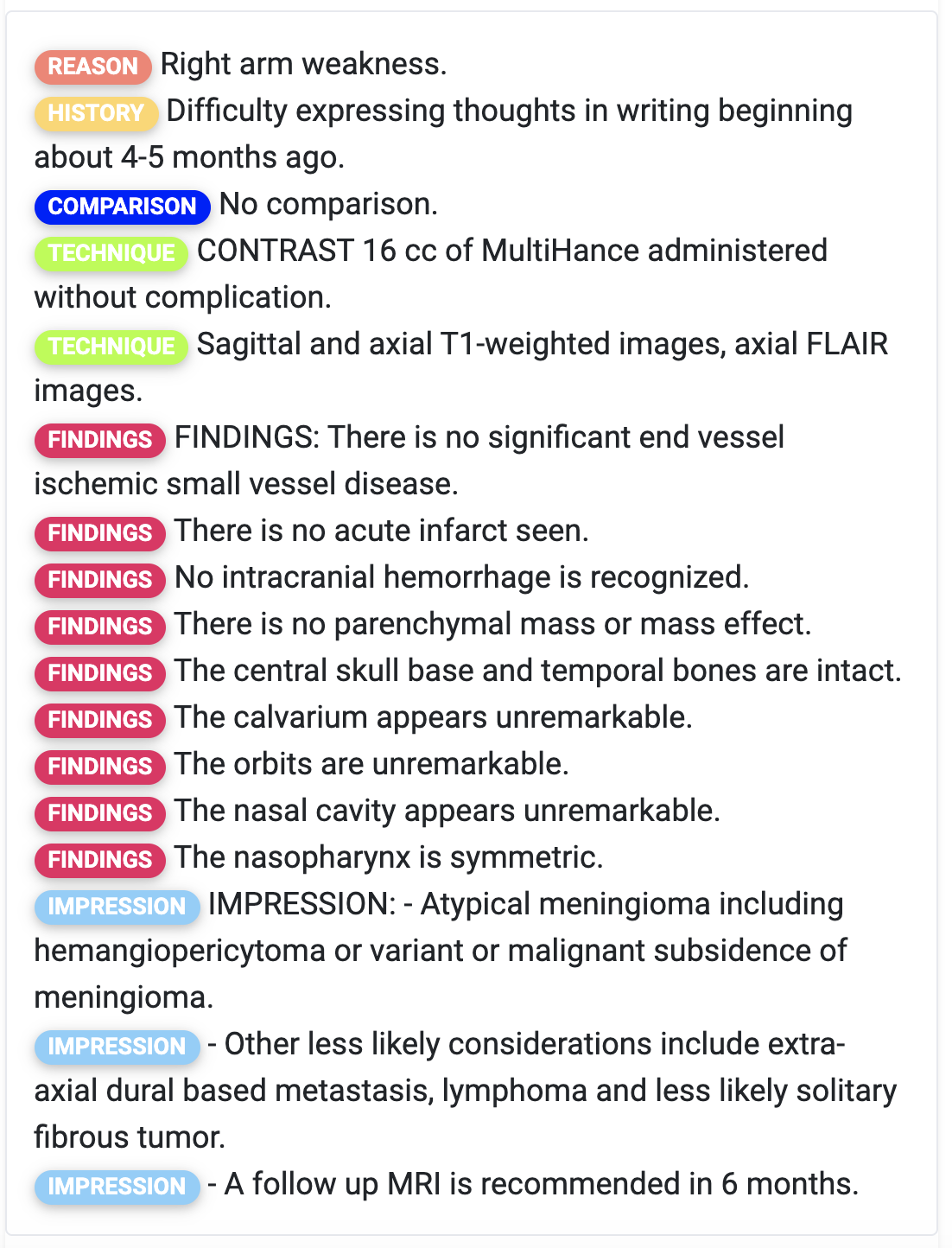}
    \caption{Snapshot of the output of our proposed model on a radiology report. Labels are shown in front of every extracted sentence.}
    \label{fig:snapshot}
\end{figure}

Developing a universal and scalable report segmenting solution is extremely challenging as most healthcare institutes use either no template or institute specific templates. Moreover, providers' style of reporting varies from one to another as sentences are written in a telegraphic format and generally do not follow English grammar rules. Nonetheless, in the case of radiology reports, the reports are often composed of similar sections, including the reason for the visit, the performed examination, a summary of observations and findings, and finally, the radiologist's impression and recommendation based on the observations.

To extract and structure patient information from notes, most clinical institutes take the approach of developing their specific set of patterns and rules to extract and label the sections within the clinical reports. This requires a substantial amount of effort for defining rules and maintaining them over time. With advancements in machine learning and NLP, researchers have more recently utilized supervised machine learning methods for automatic structuring of radiology reports \citep{apostolova2009automatic,tepper2012statistical,haug2014developing,singh2015prioritization,rosenthal-etal-2019-leveraging}. These machine learning approaches can be divided into three main themes:
1) Methods that solely rely on extracting features from the format of the text and, therefore, are biased on the specific format of the training data \citep{tepper2012statistical};
2) More recent efforts that are focused on learning to label based on the context \citep{rosenthal-etal-2019-leveraging}; and finally, 3) The hybrid approaches that combine formatting and context-driven features \citep{apostolova2009automatic}.
The two latter methods require a reasonably large amount of annotated reports and yet are not scalable solutions as they do not adequately address inter-institute variability unless model training is fine-tuned using annotated data from the target institute.

In this work, we frame the structuring of the radiology reports as a multi-class sentence classification problem. More specifically, this work presents a novel framework to identify various sections in the radiology reports and to label all sentences within the note with their corresponding section category. We propose an ensemble approach that takes advantage of formatting cues as well as context-driven features. We incorporate Recurrent Neural Networks (RNN) and sentence encoders accompanied by a set of engineered features from the reports for the task of section labeling in radiology reports. The proposed approach considers the context of the current text span and the surrounding context that helps make more accurate predictions. 

We were motivated by how a non-expert human self-teaches to perform such a task, paying attention to the context while taking formatting cues into account. We hypothesize that each of the three models learns unique and non-overlapping attributes for solving the problem at hand, and therefore, an ensemble approach seems reasonable.

In order to avoid the requirement of access to a large annotated training corpus, we follow a weak learning approach in which we automatically generate the initial training data using generic rules that are implemented using regular expressions and pattern matching.

We consider seven types of section categories and label each sentence with one of these categories. Our approach is not limited to these specific categories and it can be adapted for any template format and writing style. This is thanks to incorporating a broad set of features that are independent of physicians/institutions. Figure~\ref{fig:snapshot} depicts a snapshot of the output of our proposed model for automatic labeling of the sentences within a radiology report. The label in front of each line represents the predicted label for the following sentence.

We train and evaluate our proposed approach on a large multi-site radiology report corpus from Mass General Brigham, referred to as MGB. We demonstrate that our proposed solution significantly outperforms common existing methods for automated structuring of radiology reports~\citep{apostolova2009automatic,singh2015prioritization} as well as several baseline models. Moreover, we manually annotated 100 reports from the MIMIC-III radiology reports corpus~\citep{johnson2016mimic}, and we report performances on this dataset as well. We also make this dataset publicly available to other researchers\footnote{\url{https://doi.org/10.5281/zenodo.4074194}}. 

Our main contributions in this study are as follows:

\begin{enumerate}
    \item Investigating the importance of different types of features, including formatting and layout, as well as semantics and context in section labeling of radiology notes at the sentence level.
    \item Achieving state-of-the-art performance for automatic labeling of radiology notes with predefined section labels through an ensemble approach incorporating models that are capable of learning context and formatting features.
    \item Contributing 100 manually-annotated clinical notes with section labels at sentence-level randomly selected from the MIMIC-III corpus.
\end{enumerate}

The rest of the paper is organized as follows. In section~\ref{sec:related}, we briefly review current methods for segmenting and structuring clinical reports. Next, we describe our proposed pipeline in section~\ref{sec:methodology}. In section~\ref{sec:results}, we present and discuss our results on independent test sets, and finally, the conclusions and potential future work are presented in section~\ref{sec:conclusions}.

\section{Related Work}
\label{sec:related}

There have been numerous efforts to address the need for automatic structuring of clinical notes via section labeling, including rule-based methods, machine learning-based methods, and hybrid approaches~\citep{pomares2019current}.

\citet{taira2001automatic} proposed a rule-based system comprising a structural analyzer, lexical analyzer, parser, and a semantic interpreter to identify sections in radiology reports. \citet{denny2008development} developed a hierarchical section header terminology and a statistical model to extract section labels from sentences. RadBank was introduced by Rubin and Desser, which recognizes the structure of radiology reports and extracts the sections for indexing and search, which falls in rule-based methods~\citep{rubin2008data}. A known shortcoming of rule-based approaches is that they perform well only on reports that follow a specific template and are written following strict structures. As a result, rule-based systems require updating rules/patterns for each new dataset with new formatting and structure. Furthermore, rule-based approaches perform poorly on reports lacking a coherent structure and/or are not written using a predefined template.

Machine learning-based methods solve this problem by training models that can be applied to other datasets without substantial changes as they learn to rely on features beyond formatting and layout.
\citet{singh2015prioritization} presented a system based on the Na\"ive Bayes classifier to identify sections in radiology reports. \citep{tepper2012statistical} employ Maximum Entropy to label various sections in discharged summaries and radiology reports. \citet{cho2003automatic} proposed a hybrid system to extract and label sentences from medical documents. Their proposed system is composed of a rule-based module that detects the sections with labels and a machine learning classifier that detects the unlabeled sections. \citet{apostolova2009automatic} employed a set of rules for creating a high-confidence training set and applied Support Vector Machines (SVM) trained on additional formatting and contextual features to label the sentences from radiology and pathology reports. The main challenge in training such machine learning-based approaches is the need for a relatively large annotated training data.

To the best of our knowledge, the most recent work is proposed by \citet{rosenthal-etal-2019-leveraging} in which they present a system based on an RNN and a BERT~\citep{devlin2018bert} model for predicting sections in EHRs. They use sections from the medical literature (e.g., textbooks, journals, web content) with similar content in EHR sections.

Even though the existing methods address the problems mentioned earlier for the complex task of automatic structuring of radiology reports, an ensemble of several models is shown to yield lower generalization error, as opposed to training individual models~\citep{kotu2014predictive}.


\section{Methodology}
\label{sec:methodology}


\subsection{Approach}
In this work, we formulate the task of automated structuring of radiology reports as a supervised multi-class text classification problem. We define the label set as \textit{Reason for Visit}, \textit{History}, \textit{Comparison}, \textit{Technique}, \textit{Findings}, \textit{Impression}. Any sentence that cannot be categorized as one of the classes above is labeled as \textit{Others}. 

Suppose we have the context $C=s_1s_2...s_n$, where $s_i$ is a sentence in the radiology report. We define a mapping function $f$ that for each sentence $s_i$ from the set of sentences in the report, it maps the sentence to its associated label. The context $C$ can be the entire radiology report or a few sentences from the report. The following sections describe the details of our proposed methodology.


\subsection{Dataset}
Since we do not have access to a publicly-available dataset, we build our own training set using the radiology reports from a multi-institution clinical corpus from Mass General Brigham referred to as the MGB dataset. We randomly selected 856 radiology reports from 12 different clinical sites, i.e., Mass General Brigham. Taking the template and specific formatting/layout of the notes, we develop a weak labeler using regular expressions to detect keywords, including \textit{Findings, Impression, Technique, Comparison, Reason for Visit, History, Indications, Type}, and \textit{Procedure}. Subsequently, we consider all of the sentences between two observed keywords as the preceding section. For instance, if the keyword ``Findings" appears at position 400 and ``Impression" appears at location 700, any sentence in the range of [400, 700) is labeled as \textit{Findings}. One should note that the occurrence of all the keywords in radiology reports is not guaranteed. Therefore, we only pick the sections that appear in the report. Next, we merge \textit{Technique, Procedure}, and \textit{Type} into one category since they convey the same concept. We also combine \textit{History} and \textit{Indications} into one class. Furthermore, we manually correct the automatically assigned labels of sentences using the BRAT annotation tool~\citep{stenetorp2012brat}. Since only one human annotator corrected the labels, there is no inter-annotator agreement.
We split this dataset into three: 686 reports~(80\%) as the training set, 85 reports~(10\%) for training the ensemble model, and 85 reports~(10\%) as the test set. 

To comply with data privacy and Health Insurance Portability and Accountability Act~(HIPAA), we cannot release this dataset. Nonetheless, we randomly select a separate 100 reports from MIMIC-III corpus with the \textit{CATEGORY} code of Radiology. Subsequently, we manually annotate this dataset, similar to the way we created the MGB dataset, and we employ it as an independent test set. The annotations were performed by two of the co-authors as non-domain experts. No inter-annotator agreement was measured as there was no overlap between labeled reports by two annotators.


\begin{figure}[t]
    \centering
    \includegraphics[width=\linewidth]{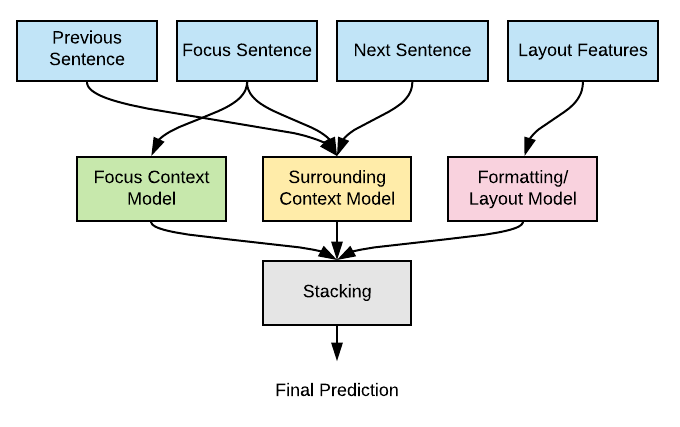}
    \caption{The ensemble model composed of the Focus Context, Surrounding Context, and Formatting/Layout models that combines the three prediction using the \textit{Stacking} method.}
    \label{fig:section_stacking}
\end{figure}

\subsection{Preprocessing}
The preprocessing includes removing special characters while keeping lowercase, uppercase, and digits from the text and replacing all other characters with space. We use Simple Sentence Segment\footnote{\url{https://github.com/noc-lab/simple_sentence_segment}} for sentence parsing. Subsequently, all of the sentences are tokenized using the SentencePiece tokenizer~\citep{kudo2018sentencepiece}. 

We utilize GloVe~\citep{pennington2014glove} word embeddings trained in-house on the entire set of radiology reports from multiple-sites~(more than two million radiology reports).
The pre-trained word embeddings are 300-dimensional. 
We also repeated our experiments by utilizing the BERT~\citep{devlin2018bert} embeddings, trained in-house on the same corpus of radiology reports, as mentioned above. Overall, the GloVe embeddings yield higher performance for the desired task compared to the BERT embeddings. Therefore, for all of the experiments, we report the performance using the GloVe embeddings.


\subsection{Model}

Figure~\ref{fig:section_stacking} demonstrates the proposed ensemble architecture. As can be seen from the figure, the three models aim to capture and encode formatting information, focus sentence context, as well as the context from the surrounding sentences of the focus sentence.

The intuition for having three models is that relying on one source, either context or format alone, is insufficient to capture all necessary text attributes for the labeling task. For example, a sentence such as ``Microlithiasis." may occur in \textit{History}, \textit{Findings} or \textit{Impression} sections and only by taking sentence context, the surrounding context, and the formatting cues altogether, one can determine the most appropriate label. 

We combine the individual models' predictions using the \textit{Stacking} method \citep{wolpert1992stacked} to derive the final prediction. The architecture of each model is discussed in detail in the following sections.

\subsubsection{Focus Context Model}
\begin{figure}[ht]
    \centering
    \includegraphics[width=0.85\linewidth]{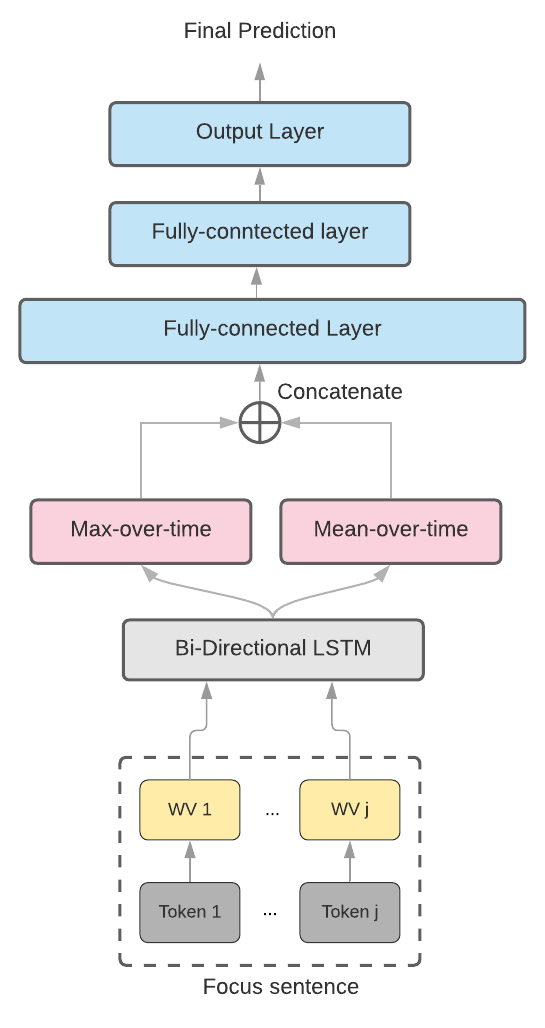}
    \caption{The network architecture of the Focus Context model.}
    \label{fig:section_single}
\end{figure}

As shown in Figure~\ref{fig:section_single}, the proposed architecture for the Focus Context model is composed of a Bi-directional Long Short-Term Memory~(LSTM) with 64 units. Subsequently, we encode the sentence using the LSTM's output sequences using max-over-time and mean-over-time pooling and concatenate these two vectors \citep{blanco2020boosting}. This approach enables us to extract meaningful features from the focus sentence context. The encoded sentence is next passed to a fully-connected layer with 100 neurons with the \textit{ReLU} activation function and a dropout value of 50\%. We stack two more fully-connected layers with sizes of 30 and 16 with the \textit{ReLU} activation functions and dropout values of 50\% and 30\%, respectively. Finally, the weights are passed to the output layer that employs a \textit{Softmax} activation function to make the final prediction. 

\begin{figure*}[t]
    \centering
    \includegraphics[width=0.9\linewidth]{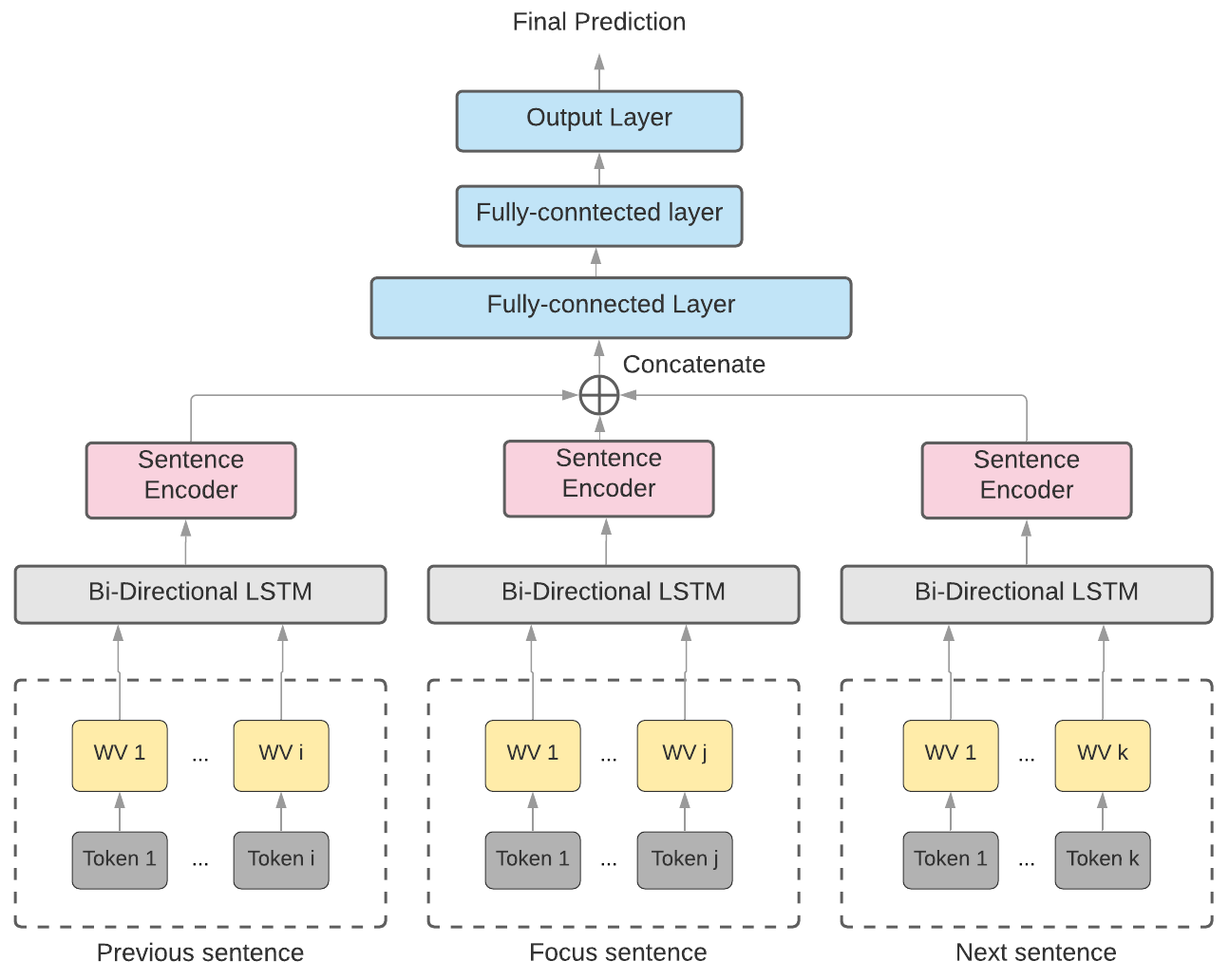}
    \caption{The network architecture of the Surrounding Context model.}
    \label{fig:architecture}
\end{figure*}

\subsubsection{Surrounding Context Model}
Figure~\ref{fig:architecture} demonstrates the proposed architecture for the Surrounding Context model. The surrounding context is defined as the sentence immediately before and the sentence immediately after the focus sentence. The most efficient size of the surrounding context can be determined through hyper-parameter tuning, which is beyond the scope of this work and is considered for future work. Each sentence is fed into a Bi-directional LSTM layer. The LSTM layer for the focus sentence comprises 64 units, whereas the LSTM layers of surrounding sentences have 16 units.
Next, each Bi-LSTM layer's output sequence is fed into a max-over-time pooling layer to encode the sequence. The three sentence encoders' outputs are concatenated and passed into a fully-connected layer with 50 neurons and \textit{ReLU} activation function. This layer is followed by a Dropout layer with a value of 50\%. The weights are passed to a fully-connected layer with ten neurons and a dropout value of 30\%. Subsequently, the output is fed into a second fully-connected layer with seven neurons and the \textit{Softmax} activation function to obtain the final prediction. In cases where the focus sentence appears at either the beginning or end of a report, we use an empty string for the sentence before or after.

\subsubsection{Formatting/Layout Model}
We propose a third model to learn formatting/layout related features using neural networks. Motivated by a prior work \citep{apostolova2009automatic}, we define 17 features that are described as follows:

\begin{enumerate}
    \item Number of uppercase, lowercase, and digits in the sentence~(three features).
    \item Normalized relational position of the focus sentence to each section headers by searching keywords such as reason, history/indications, procedure/technique, comparison, findings, and impression~(six features).
    \item If the last character of the previous sentence, the current sentence, and the next sentence is either period or colon~(six features).
    \item Normalized position of the current sentence in the report~(one feature).
    \item If the first token in the sentence is uppercase or not~(one feature).
\end{enumerate}

These features are utilized as input to a neural network with a stack of three fully-connected layers with 100, 16, and seven neurons. We add the \textit{ReLU} activation functions for the first two layers and the \textit{Softmax} function for the last layer. The first two layers are followed by dropout layers with values of 50\%.

\subsubsection{Ensemble: Stacking}
As the last step, we train a Logistic Regression (LR)-based ensemble model using the three models described in the previous sections and using a holdout stacking set. 
We start making predictions using the three models on the holdout set, and we train an LR classifier on their predicted probabilities using Equation~\ref{eq:logistic},

\begin{equation}
\label{eq:logistic}
    p(y=1)= \sigma(w^Tx+b)
\end{equation}

\noindent
where $w$ and $b$ are parameters to learn from data, and $\sigma$ is the Sigmoid function. We perform ``one-versus-rest" for multi-class classification.
The trained classifier can be utilized for making accurate predictions on the test set. 

\subsection{Experimental Setup}
We implement four baseline models to compare with our proposed model. The first baseline is a rule-based model using the regular expressions specifically assembled based on the format of radiology reports from the MGB dataset. We refer to this model as the \textit{MGB Rule-based} model. The second baseline is also a rule-based model composed of rules designed specifically for the MIMIC-III dataset. We refer to this model as the \textit{MIMIC Rule-based} model. The third baseline model is a neural network consist of similar architecture to ours, but instead of stacking, we concatenate the outputs and pass it to a fully-connected layer. We refer to this model as the \textit{Merged} model. 
We also compare our proposed ensemble model with a Linear SVM model with ``balanced" class weights, trained on preprocessed sentences in the form of uni-gram TFIDF vectors.


\begin{table*}[t]
\centering
\begin{tabular}{l|cc|cc}
\toprule
{}             & \multicolumn{2}{c}{\textbf{MGB-test}} & \multicolumn{2}{c} {\textbf{MIMIC-III}} \\ 
\textbf{Model} & \textbf{Accuracy} & \textbf{m-F1} & \textbf{Accuracy} & \textbf{m-F1} \\ \hline
\midrule
MGB Rule-based                     & 62.7\%   & 47.3\% & 29.7\%   & 23.6\%    \\
MIMIC Rule-based                  & 57.5\%   & 31.0\% & 33.4\%   & 30.7\%    \\
Linear SVM                         & 89.4\%   & 82.2\% & 66.2\%   & 63.3\%    \\ \hline
\citet{apostolova2009automatic}   & 90.3\%   & 84.3\% & 72.1\%   & 69.4\%    \\
\citet{singh2015prioritization}   & 85.9\%   & 74.7\% & 68.8\%   & 63.6\%    \\ \hline
Formatting/Layout                 & 92.3\%   & 75.2\% & 42.1\%   & 40.6\%    \\
Focus Context                     & 89.4\%   & 74.3\% & 62.0\%   & 55.3\%    \\
Surrounding Context               & 93.7\%   & 88.8\% & 71.2\%   & 67.9\%    \\
Merged Ensemble                   & 94.3\%   & 89.3\% & 73.3\%   & 69.2\%    \\
Stacking Ensemble                 & \textbf{97.1}\%   & \textbf{93.7} & \textbf{77.5}\%   & \textbf{74.0}\% \\
\bottomrule
\end{tabular}
\caption{Comparison of the results of various models on the MGB-test set and 100 MIMIC-III notes. m-F1 stands for macro F1 score across seven classes.}
\label{tab:results}
\end{table*}

Most prior approaches utilize specific labeling schema that differ from ours and the corresponding labeled datasets are not publicly available~\citep{cho2003automatic,rubin2008data,apostolova2009automatic,singh2015prioritization}. As a result, we cannot provide a fair comparison of our proposed model with such approaches. Moreover, some studies employ external data sources during training, e.g., journals and textbooks~\citep{rosenthal-etal-2019-leveraging}, which is also not compatible with the radiology report labeling schema, and restricts us from comparing our model with their work. Nevertheless, we implement the two existing methods presented by \citet{apostolova2009automatic} and \citet{singh2015prioritization}, which label sections in radiology reports. Since we did not have access to their code, we tried to replicate their methods to the best of our knowledge and understanding.

We implement our proposed model using Keras~\footnote{\url{https://keras.io}}. We utilize Adam optimizer with a learning rate of 0.001 and Categorical Cross-Entropy loss. 
We split the training set into two sets: 90\% for training and 10\% as the validation set. We use early stopping by picking the best validation accuracy value among 30 epochs for the models with the patience value of five. We also set the patience value to 200 among 600 training epochs for the Layout model.

We run our experiments on an \textit{Amazon c5.18xlarge EC2} instance\footnote{\url{https://aws.amazon.com/ec2/instance-types}}. The average running time for the focus context, surrounding context, Formatting/Layout, and Merged models are roughly 80, 70, 60, and 60 minutes, respectively. 

\section{Results and Discussion}
\label{sec:results}

\subsection{Model Comparison}

We compare our proposed Stacking Ensemble model with several prior work as described above. We also report the performance of individual models used in our Stacking Ensemble model to investigate the importance of each model independently. Table~\ref{tab:results} summarizes the performance of different approaches in terms of accuracy and macro F1 on the MGB-test set as well as 100 MIMIC-III notes.  

It can be observed that, overall, our proposed Stacking Ensemble model outperforms all other approaches on both test sets. By comparing the performance of the three models composing our proposed ensemble model, we observe that the Surrounding Context model achieves the highest performance among three, emphasizing the importance of the surrounding context in such a labeling task. Furthermore, it can be observed that the Formatting/Layout model performs worse on MIMIC-III set than the MGB-test set. This could be because reports from the MGB set are structured more consistently than MIMIC-III notes. In other words, MIMIC-III notes are not prepared using a specific and consistent template.

Another observation is that the rule-based models, i.e., MGB Rule-based and MIMIC Rule-based, perform poorly compared to machine learning-based approaches even though they are tailored specifically based on the corresponding reports' format and structure. Moreover, we observe that the MIMIC Rule-based model yields lower accuracy than MGB Rule-based model on the MGB-test set and vice versa. This confirms that the performance of rule-based approaches significantly varies across different datasets, and overall, rule-based approaches suffer from generalization and scaling.

Finally, the proposed Stacking Ensemble model yields lower performance on MIMIC-III test set compared to the MGB-test set. This could be because there are significant differences between the two sets of radiology reports in terms of content and format: MGB-test set notes are from inpatient and outpatient care and in general, follow a consistent format; however, MIMIC-III reports are discharge notes from the Emergency Department lacking a consistent structure.

To evaluate the sensitivity of our proposed model to a particular split of the data, we perform 10-fold cross-validation on the training set (i.e., split between 90\% training and 10\% validation). The $mean\pm std$ of accuracy and macro F1 across 10-folds are $97.0\%\pm0.2\%$ and $93.0\%\pm 0.2\%$, respectively.





\begin{figure}[thb]
    \centering
    \includegraphics[width=\linewidth]{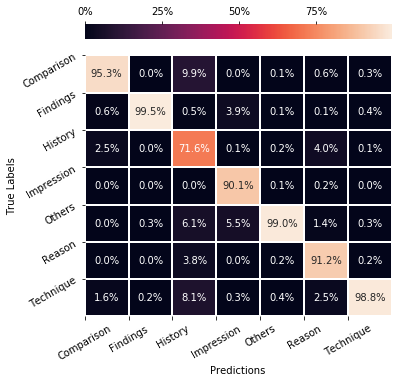}
    \caption{Confusion matrix showing the percentages of true and mislabeled predictions on the MGB-test set.}
    \label{fig:confusion_test}
\end{figure}

\subsection{Error Analysis}

\begin{figure}[thb]
    \centering
    \includegraphics[width=\linewidth]{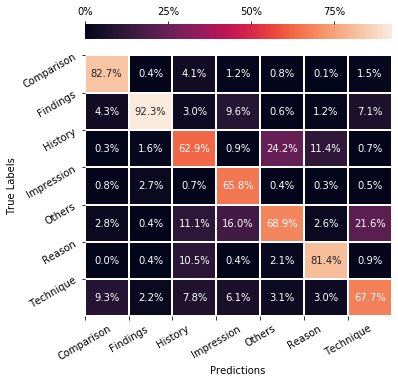}
    \caption{Confusion matrix showing the percentages of correct and incorrect predictions on MIMIC-III set.}
    \label{fig:confusion_mimic}
\end{figure}

We further investigate the performance of the Stacking Ensemble model for each class label separately. Figures~\ref{fig:confusion_test} and \ref{fig:confusion_mimic} depict the confusion matrices between the predictions and actual labels from each class for MGB-test set and MIMIC-III notes, respectively. It can be observed that among all classes, ``History" is the most challenging, and it is occasionally misclassified as ``Comparison" in the case of MGB reports, and with ``Others" and ``Reason" classes in the case of MIMIC-III reports. We consider two possible reasons for this: 1) the similarity of the context between ``History" and the other classes as mentioned earlier; and 2) the adjacency of these sections within the radiology reports.

\subsection{Analysis of Stacking Ensemble Input}
\begin{figure}[htbp]
    \centering
    \includegraphics[width=0.9\linewidth]{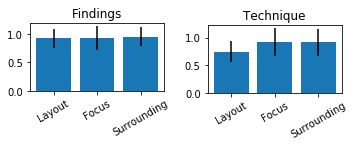}
    \caption{Comparison of the errors of inputs to the Stacking Ensemble model for the Findings and Technique classes on the MGB test set.}
    \label{fig:technique_comparison}
\end{figure}

To further investigate each type of model's importance in the final ensemble decision, we analyze the weights resulting from the ensemble. We observe the different distribution of weights for different label types. For example, weights are equally distributed among three models for ``Finding" and ``Impression" sections. On the other hand, we observe unbalanced weight distribution for ``Technique" and ``Comparison" classes. Figure~\ref{fig:technique_comparison} shows the mean of weights for the ``Findings" and ``Technique" classes on the MGB-test set. It can be seen that all the models are equally important for the ``Findings" class, whereas, for the ``Technique" class, there is less emphasis on the Formatting/Layout model than the Focus Context and Surrounding Context models.

\subsection{Fine-tuning the Stacking Model}
As can be seen from Table \ref{tab:results}, the proposed ensemble model trained on MGB data does not perform as well on the MIMIC-III set. We try to improve the performance of the proposed ensemble model on the MIMIC-III set by fine-tuning the ensemble part on a MIMIC-III data subset. We split the MIMIC-III data into 20\% for fine-tuning and 80\% for testing. Table~\ref{tab:finetuning_results} demonstrates the results of running the Stacking Ensemble model on 80\% of the MIMIC-III data with and without fine-tuning. As can be seen from the table, we can obtain a 5.5\% increase in accuracy score and a 6.9\% increase in macro F1 score. This is achieved by only fine-tuning the ensemble step using a small subset of the MIMIC-III data, while the individual models are still trained on the MGB data.

\begin{table}[htbp]
\centering
\begin{tabular}{l|c|c}
   \textbf{Model Type}                & \textbf{Accuracy} & \textbf{m-F1} \\ \hline
Without fine-tuning & 76.3\%   & 73.9\%    \\
With fine-tuning    & \textbf{81.8}\%   & \textbf{80.8}\%   
\end{tabular}
\caption{Comparison between the performance of Stacking Ensemble model with and without fine-tuning on MIMIC-III data.}
\label{tab:finetuning_results}
\end{table}

To show that the results are not sensitive to any specific split of data, we perform five-fold cross-validation on the MIMIC-III reports by utilizing 20\% of reports for training the Logistic Regression classifier, and 80\% for testing. We obtain a mean value of 81.5\% accuracy with a standard deviation of 0.7\% and a mean value of 80.4\% macro F1 score with a standard deviation of 0.9\%, which shows the insensitivity to the specific split of data.

A known shortcoming of our proposed approach is the sensitivity to the accuracy of the sentence segmentation. Poor sentence parsing results in misslabeling, specifically, if error in sentence parsing results in combining sentences belonging to two different sections. To address this issue, we are currently working on training a clinical note-specific sentence parsing algorithm. 

\section{Conclusions and Future Work}
\label{sec:conclusions}

In this work, we propose an ensemble approach for automatically labeling sentences in radiology reports with section labels. Through the proposed ensemble approach, we achieve the state-of-the-art performance of 97.1\% on a relatively sizeable multi-site test set from Mass General Brigham. Our proposed ensemble method is composed of three parallel models that capture various structural and contextual attributes of radiology reports, including formatting/layout, focus context, and the surrounding context. Furthermore, We compared our proposed ensemble model against each of its components and concluded that the combination of all models is more accurate than any individual model. 

As future work, we plan to incorporate performance calibration in our ensemble model. It adds the importance of individual models to the ensemble model and enables us to obtain higher performance for unseen data. We also plan to extend this work to other types of reports, i.e., pathology reports and discharge summaries. Another potential future work is to add Conditional Random Field~(CRF) to our proposed model.



\bibliographystyle{acl_natbib}
\bibliography{emnlp2020}








\end{document}